# *Contradiction Detection in Persian Text*

Zeinab Rahimi and Mehrnoush ShamsFard

*NLP lab, Shahid Beheshti University, Tehran, Iran*

*email*: *{z_rahimi@sbu.ac.ir, m-shams@sbu.ac.ir}*

**Abstract**

Detection of semantic contradictory sentences is one of the most challenging and fundamental issues for NLP applications such as recognition of textual entailments. Contradiction in this study includes different types of semantic confrontation, such as conflict and antonymy. Due to lack of sufficient data to apply precise machine learning and specifically deep learning methods to Persian and other low resource languages, rule-based approaches that can function similarly to these systems will be of a great interest. Also recently, emergence of new methods such as transfer learning, has opened up the possibility of deep learning for low-resource languages. Considering two above points, in this study, along with a simple rule-base baseline, a novel rule-base system for identifying semantic contradiction along with a Bert base deep contradiction detection system for Persian texts have been introduced. The rule base system has used frequent rule mining method to extract appropriate contradiction rules using a development set. Extracted rules are tested for different categories of contradictory sentences. In this system the maximum f-measure among contradiction categories is obtained for negation about 90% and the average F-measure of system for all classes is about 76% which outperforms other algorithms on Persian texts. On the other hand, because of medium performance of rule base system for some categories of contradiction, we use a Bert base deep learning system using our translated dataset; with average F-measure of 73. Our hybrid system has f-measure of about 80.

## 1. Introduction

Contradiction detection is a fundamental task in text understanding and has many possible applications especially in textual inference and sentiment analysis. The contradiction relation represents the semantic confrontation between two entities and in general includes all antonymy, negation and other semantic confrontation such as world knowledge, lexical, numeric and structural. The words "contradiction",

"contrast", "confrontation", and "conflict" in this domain, are sometimes used interchangeably, despite the differences they have.

Contradiction usually occurs in two levels of word and structure. In word level, semantic contradiction includes the following different types (Safavi, 2014):

1) Negation: negation refers to the cases in which the negation of one is proof of the other, such as "man (مرد)" and "woman (زن)", "on (روشن)" and "off (خاموش)", "open (باز)" and "closed (بسته)". This kind of contradiction is called complementary contradiction.

2) Gradable adjectives: This type refers to the words in a spectrum of values of a specific attribute, such as: "cold-cool-warm-hot (سرد-خنک-گرم-داغ)" and neglecting one is not a proof of the other. Word pairs such as: "cold and hot (سرد و داغ)", "short and long (کوتاه و بلند)", "big and small (بزرگ و کوچک)", "old and young (پیر و جوان)", "ugly and beautiful (زشت و زیبا)" are referred to as **scalar contrasts**, so if we say that "It is not cold", it does not mean that "it is necessarily warm". Or, if we say that "she is not ugly", it does not mean that "she is beautiful".

3) Double-confrontation: In this type, there is a two-way relationship between two words. Like a couple (wife and husband- زن و شوهر) or buyer and seller (خریدار و فروشنده). This means that if A bought something from B, then it's likely that B has sold it to A. Or if A is the husband of B, then surely B is A's wife.

4) Lexical opposition: This type is made up using a negative affix. Such as knowingly and <u>un</u>knowingly (آگاه و ناآگاه), polite and <u>im</u>polite (باادب و بی ادب) or safe and <u>un</u>safe. In this type, the proof of one is usually the negation of the other, and vice versa, for example, "he is not a polite man." That is, "He is impolite".

5) Directional opposition: In this case a point is considered and the pair of words are measured according to that point. For example, "going (رفتن)" is essentially moving away from that point, and "coming (آمدن)" is getting closer to that point. Thus, "coming and going (رفتن و آمدن)", "up and down (بالا و پایین)", "bringing and taking (آوردن و بردن)", "backward and forward (جلو و عقب)", "left and right (چپ و راست)" and "north and south (شمال و جنوب)" are examples of words that have directional opposition.

These types are commonly referred to words without considering their contexts, but in most of the times, the focus is on the contrast detection between sentences. That is, in the area of interest we will be confronted with different types of contradictions between sentences.

As mentioned before, the **opposition** (or antonymy), predominantly stands between two words or phrases in which the negation of one is a further proof for the other, such as "on" and "off", or words or phrases which are created with negation affixes such as "accurate" and "inaccurate", while **contradiction** is mainly between two sentences or a pieces of text, in which two sentences are extremely unlikely to be true simultaneously, such as "Mary was killed yesterday" and "Mary is eating lunch at the restaurant now."

So, considering the types of confrontation that occur in textual entailments, we can have another categorization: (Marneffe et al., 2008)

1) Antonym words: In this category, antonym words appear in similar semantic roles in two sentences. For example, two sentences "soup is *hot* (سوپ داغ است)" and " soup is *cold* (سوپ سرد است)" are opposites. "Capital punishment is a *catalyst* for more crime." and "Capital punishment is a *deterrent* to crime." are another examples for opposite sentences with antonym words.

2) Negation: In this category, two sentences are contradictory with negativity of a position using the negation sign. Like "Saman went to school (سامان به مدرسه رفت)" and "Saman did not go to school (سامان به مدرسه نرفت)". The same situation is held between "A closely-divided Supreme Court said that judges and *not judges* must impose a death sentence." and "The Supreme Court ruled that *only judges* can impose the death sentence".

3) Numerical: In this category, contradiction occurs due to the numerical contrast in two sentences. Like " more than 100 people were killed in the war (در جنگ بیش از ۱۰۰ نفر کشته شدند)" and "50 people were killed in the war. (در جنگ ۵۰ نفر کشته شدند)". Or "The tragedy of the explosion in the city that *killed more than 50 civilians* has presented government with a dilemma." and "An investigation into the strike in the *city found 28 confirmed dead* so far."

4) Factive: In this category, contradiction occurs when manner of expressing of an event or action in the first sentence makes an assumption in mind, whose opposite is being stated in the second sentence. For example, the sentences "The thieves did not intend to enter the bank (دزدها قصد نداشتند که وارد بانک شوند)," and "thieves entered the bank. (دزدها وارد بانک شدند)" Or "Prime Minister says he will not be swayed by a warning that country faces more terrorist attacks *unless it withdraws* its troops from north." and "Country *withdraws* from north."

5) Structural: In this category, conflict has been made due to changes in the components of a relationship. That is, the semantic structure of the relation of the first sentence has changed in the second sentence and the meaning has been changed, but syntactically and without considering the first sentence, the second sentence may be correct. Like "Trump won Clinton in elections (ترامپ کلینتون در انتخابات)" and "Clinton won Trump in elections (در انتخابات کلینتون را برد ترامپ را برد)" Or "The Channel Tunnel stretches from *England to France*. It is the second longest rail tunnel in the world, the longest being a tunnel in Japan." and "The Channel Tunnel connects *France and Japan.*"

6) Lexical: In this category, a word or phrase is opposite with the word or phrase in the second sentence while these phrases are not necessarily opposite, but they are contradictory in the context of these two sentences. Like "Mr. Rahimi educates his children in a modern way (آقای رحیمی فرزندانش را به روش مدرن تربیت می‌کند.)" and "Mr. Rahimi forces his son to study (آقای رحیمی پسرش را مجبور می‌کند درس بخواند.)". Or "The Canadian Parliament's Ethics Commission *said* the former immigration minister, Judy Sgro*, did nothing wrong* and her staff had put her in a conflict of interest." and "The Canadian Parliament's Ethics Commission will *accuse* Judy Sgro."

7) World Knowledge: Two sentences contradict from the background knowledge that they convey. Such as the " AmirKabir University was founded in 1337. (دانشگاه امیرکبیر در سال ۳۷ تاسیس شد)" and "My father was student in AmirKabir University in 1335 (پدر من در سال ۳۵ دانشجوی دانشگاه امیرکبیر بوده است). " Or "One of the first Microsoft branches outside the USA, was founded in 1989." and "Microsoft was established in 1989."

Among above categories, the first, second and third categories have been considered in most papers and the last four categories are less studied in researches up to authors' knowledge.

Although the existing methods would be useful to recognize most of the word-level oppositions, but because of the differences in structural patterns of oppositions in Persian language, they must be changed and adopted and will be considered within the scope of current research. Some examples of these differences are given below:

- When using quantifier to negate sentences such as "no one came," in English we have negative quantifier and positive verb while in the same sentence in Persian ("هیچ کس نیامد") negative verb and negative quantifier are used.

- When using negatives adverbs like "never (هرگز)" we have the same pattern: negative adverb and positive verb in English and negative verb and negative adverb in Persian.

- When using the terms "هم (neither)" in "Mary does not like this food. Neither do I. (مریم این غذا را دوست ندارد. من هم دوست ندارم)", In Persian, negative verb is used in both sentences, while in English the verbs are positive. In addition, in Persian the adverb ("هم") is the same for both positive and negative states, while in English "neither" and "either" are used for negative and positive sentences respectively.

So as the novelty of this paper: 1) considering the different patterns of contradictions in Persian, we inspect the ways contradictions occur across Persian texts and describe two rule-base systems for automatically detecting such constructions. And 2) To cover some contradictory cases which are not properly solvable through rules, we prepare a training dataset for presenting a Bert base deep learning system too.

The rest of the paper is organized as follows. In section 2 related researches are reviewed. In Section 3, the proposed methods are fully explained, in section 4 evaluation details are stated and finally in section 5, conclusion and further works are indicated.

## 2. Related Work

There are few work done with the raw subject of automatic contradiction detection. Instead there are some contradiction detection works embedded in various applications of the natural language processing, such as sentiment analysis and textual entailment. On the other hand, if we want to classify the proposed methods in related researches, we can define three categories:

1- Rule-based methods e.g. (Harabagiu et, al., 2006), (Marneffe et, al., 2008) and (Blanco and Moldavan, 2011), (Asmi and Ishi, 2012) and (Quang et al., 2013)
2- Machine learning and deep learning methods such as (Rocktaschel et, al., 2016), (Wang and Jiang, 2016), (Khandelwal and Sawant, 2019) and (Sifa et al., 2019).
3- Other approaches such as (Shi et al., 2012), (Vargas et, al., 2017) and (Li et al., 2017).

In this section we introduce the related researches and their details.

As Marneffe (2008) says, Harabagiu and collegues (2006) provide the first empirical results for contradiction detection. In (Harabagiu et, al., 2006), three categories of negation, antonyms and semantic information related to contrast are considered. It has been argued that this information can be used to detect incompatible information (such as two conflicting answers to a question in a Q&A system) or to identify compatible information (semantic similarity, redundancy, and textual entailment). The goal is to find contradictory information in the text, and a framework for identifying contradictions is proposed that addresses negation, antonymy and contradiction. To detect negation, two approaches are considered: (1) Direct negation (not, negative quantifiers (no, no one, nothing, and negative adverbs like never) and (2) Indirect negation (verbs like deny, fail, refuse, prepositions like without, weak quantifiers like few, any, and some, and cases of traditional negative polarity like any more)

As it was discussed, Marneffe classifies the contradiction into the categories of antonyms, negation, numerical, Factive, structural, lexical, and world knowledge, and argues that all these can be again categorized into two categories: (1) negation and antonyms; The mismatch between date or number, and (2) the contradictions derived from the use of modal words, factive, lexical and knowledge-based contradictions. In (Marneffe et, al., 2008), the first category is considered. And the following steps are taken:

-Language analysis: Representation of a language to display text content
-Graph alignment: Using Stanford parser, the text and the hypothesis are converted to dependency graphs, and would be aligned.
- Filter non-coreferenced events
- Extraction of contradictory features (including polarity, numbers and date contradiction, structure, …)

Blanco and Moldavan (2011), analyse some cases about identifying negations in texts and their scope. The main idea is that in cases involving the discovery of negation, two issues are important: (1) scope and (2) focus. Scope is part of the meaning that is negated and is a territory for affected area. With regard to that limitation, the negated or non-negated parts of the sentence is determined. For example, the sentence "All vegetarians do not eat meat" goes back to the entire vegetarian, but the sentence "All plants are not eaten by vegetarians" does not mean that vegetarians just eat plants. The second issue that is focus, is about the part of sentence which has the main attention and the way it can be negated. For example, in the sentence "That land was not large, it was huge," the magnitude of the land is worth considering. This paper proposes rules for different categories of negation add them to the representation of a negated sentence.

In (Sankara and Mehta, 2011) a fuzzy rule based algorithm is proposed. The steps of this algorithm are:
- Preprocessing and removing additional characters and reformatting abbreviations, numbers and characters.
- Specifying the keywords of the document and initializing zero to 3 classes of aligned, conflicting and unrelated cases.
- Comparing keywords together and increase the value of each of the three above variables in case of occurrence
- Calculating the final value of variables and making initial decision
- Calculating the matching ratio (the number of conflicting words out of the sum of words aligned and non-correlated) and make final decision based on the thresholds and the obtained ratio.

In (Asmi and Ishi, 2012), the identification of negation in a sentiment analysis system has been investigated. In this regard, sentences that are negated are identified using a dependency parser, and the polarity of the negated sentences is calculated using a set of rules extracted from the senti-wordnet.

In (Shi et al., 2012), the problem of knowledge scarcity is presented as the problem of contradiction extraction. In this regard, authors using a web query, measure the frequency of phrases having non-matching relationship, and analyse the adaptation degree of these non-matches to the existence or non-existence of a contradiction.

As we mentioned, some contradiction detection work embedded in some applications of the natural language processing, such textual entailment. In textual entailment task, we have two text fragments called 'Text' and 'Hypothesis' and the goal is determining whether the meaning of the hypothesis is entailed (can be inferred) from the text or the pair are contradictory or neutral. In (Quang et al., 2013), combination of shallow semantic representations using semantic role labelling and binary relations extracted from sentences using a rule-based method has been used. Initially, after the syntax analysis using coreNLP[1] library, the Senna tool is used to label semantic roles. Then, using the REVERB tool, binary relations are extracted from sentences. This tool takes a piece of text with POS tags as input and creates output triples in the form (argument 1, relationship, argument 2). In the process of detecting the contradiction, two steps are taken to detect the conflicts of the frames and to detect the contrast of the relationships. In the first step, the verbs of the text and the hypothesis are compared using sources such as VerbNet and VerbOcean and are placed in one of the categories of matching, conflict or non-related. Also, semantic frames of text and hypothesis are scored by using a conflict function according to the inconsistency of their events. On the other hand, some extracted relations from the text and the hypothesis are compared and decisions are made.

In (Vargas et, al., 2017) a sentiment based contradiction detection system has been proposed which assumed oppositions as antonymy or contradiction, based on how many topics or attributes a sentence refer. For each input sentence pairs, topics are extracted and then sentence polarity is measured for each topic. If polarity for a single topic differs in two sentences and their calculated similarity is less than a threshold, the sentence pair are tagged as contradictory.

---

[1] https://stanfordnlp.github.io/CoreNLP/

In (Li et al., 2017) a contradiction specific word embedding (CWE) is constructed. In a way that with using antonym and negation based contrasts, some artificial contradictory sentences are made and used as training set to form a contradiction word embedding space. After this stage, this embedding is used to detect contradictions.

Except for the above papers, due to this issue that textual entailment dataset mostly has contradiction tags too (mainly after publishing of SNLI corpus -RTE corpus of Stanford university which is introduced in section 2.2- in 2015,), some of RTE[2] methods which use classification algorithms and especially deep learning methods, can be considered as related works. Here we introduce some of the most important systems in this group.

(Rocktaschel et, al., 2016) propose a neural model that reads two sentences to determine entailment using long short-term memory units (LSTM). And extend the model with a word-by-word neural attention mechanism that encourages reasoning over entailments of pairs of words and phrases. The paper did not directly report the system performance for contradiction class, but because of the novel idea and good overall performance of system, the proposed network is basis for many next researches.

In (Wang and Jiang, 2016) a special long short-term memory (LSTM) architecture for NLI[3] has been proposed. Their model builds on top of a neural attention model for NLI but is based on a different idea. Instead of deriving sentence embeddings for the text and the hypothesis to be used for classification uses a match-LSTM to perform word-by-word matching of the hypothesis with the text. They claim this LSTM is able to place more emphasis on important word-level matching results. The authors have reported promising result in both entailment and contradiction classification tasks.

Lingam and colleagues (Lingam et al., 2018) propose an approach for detecting three different types of contradiction: negation, antonyms and numeric mismatch. They derive several linguistic features from text and use it in a classification framework for detecting contradictions with artificial neural networks and deep learning techniques such as Long short-term memory (LSTM) and Global Vectors for Word Representation (GloVe).

Research of Tawfik and Speruit (2018) introduces an automated two-phase contradiction detection model that integrates semantic properties as input features to a Learning-to-Rank framework, to identify key findings of a research article. In fact, identifying inconsistencies in text are done through a two-phase algorithm: claim retrieval and claim assertion. During the first phase, they identify potential sentences relevant to the query and in the claim assertion phase, they evaluate whether sentences infer text entailment or contradiction. It also relies on negation, antonyms and similarity measures to detect contradictions between findings.

In (Khandelwal and Sawant, 2019), the decision choices for negation detection and scope resolution involved with are explored using BERT (a popular transfer

---

[2] recognizing textual entailment
[3] Natural Language Inference

learning model) on 3 corpora: The BioScope Corpus, the Sherlock dataset, and the SFU Review Corpus and then, report state-of-the-art results for scope resolution across all 3 datasets

The research of (Sifa et al., 2019) examine a set of baseline methods for the contradiction detection task on German text. For this purpose, the well-known Stanford Natural Language Inference (SNLI) data set (110,000 sentence pairs) is machine-translated from English to German. they train and evaluate four classifiers on both the original and the translated data, using state-of-the art textual data representations. The main contribution is the first large-scale assessment for this problem in German, and a validation of machine translation as a data generation method.

After we have seen the contradiction types in introduction section and the algorithms proposed to solve the problem in section 2, it seems that we can classify algorithms for different types of contradiction as bellow. We performed experiments in section 4 that somehow prove this algorithm dedication.

Table 1. algorithms for to different types of contradiction

| Using ontologies and knowledge bases | Opposite words ratio (from what adopted) | Using conflict feature set (TMP,LOC, sentiment ...) | Semantic role labelling | Sentiment analysis | Neural network, deep learning and word embedding | Rule-based | Contradiction type/Algorithm |
|---|---|---|---|---|---|---|---|
|   | * |   |   | * | * | * | **antonymy** |
|   | * |   |   | * | * | * | **negation** |
|   |   | * |   |   | * | * | **numeric** |
| * |   |   |   |   | * |   | **Factive** |
|   |   |   | * |   | * |   | **structural** |
| * |   | * |   |   | * | * | **lexical** |
| * |   |   |   |   | * | * | **World knowledge** |

There is not much work on Persian language in the field of automatic recognition of contradiction (except for linguistic researches). One of the researches has been devoted to the negation problem in the context of opinion mining. (Noferesti and ShamsFard, 2016)

In the work of (Khodadadi et al., 2015), discourse signs in Persian textual corpus were used to recognize the contradiction and a machine learning system was trained. This system determines contradiction relations in one sentence and uses discourse signs such as "but".

Recently (Amirkhani et al., 2020) present a new dataset for the NLI task in the Persian language (FarsTail) which contains contradiction data too. This research also presents the results of traditional and state-of-the-art methods on FarsTail including different embedding methods such as word2vec, fastText, ELMo, BERT, and LASER, as well as different modelling approaches such as DecompAtt, ESIM, HBMP, ULMFiT, and cross-lingual transfer approach to provide a solid baseline for the future research.

As we mentioned at the end of introduction section, in this paper, we present two rule base contradiction detection methods and evaluate them on our created Persian datasets. Also due to emergence of Google Bert and its great performance on similar tasks and also on NLI task in other languages, we implemented a deep learning Bert based system on our data.

### *Related Datasets*

There are few manually tagged datasets for contradiction detection task. Some of them are following.

- The RTE competition datasets

These datasets are prepared for RTE competitions from 2005 to 2010 yearly and include training, testing and development subsets. The competitions have continued under TAC and SemEval competitions since then. RTE1 to RTE6 datasets in total have about 35000 sentence pairs which are manually labelled with three categories: "Yes(entailment), NO(contradiction), and unknown".

- SICK corpus

This collection consists of 10,000 English sentence pairs from two sources: ImageFlickr and SemEval2015 video description, which are manually labelled with three categories: "Entailment, Contradiction, and Neutral".

- Stanford University SNLI corpus

This corpus is a collection of 570k human-written English sentence pairs, which is manually labelled with three categories: "Entailment, Contradiction, and Neutral".

-The Multi-Genre Natural Language Inference (MultiNLI[4]) corpus is a crowd-sourced collection of 433k sentence pairs annotated with textual entailment information. The corpus is modelled on the SNLI corpus, but differs in in some ways; it covers a range of genres of spoken and written text, and also it supports a distinctive cross-genre generalization evaluation.

-ES-C: A contradiction annotated dataset in Spanish within the news domain which sentences are classified as compatible, contradictory, or unrelated information. The dataset consists of 7403 news items, of which 2431 contain Compatible headline–body news items, 2473 contain Contradictory headline–body news items, and 2499 are Unrelated headline–body news items. Presently, four different types of contradictions are covered in the contradiction examples: negation, antonyms, numerical, and structural. (Sepúlveda-Torres et al., 2021)

-FarsTail[5]: includes 10,367 samples which are provided in both the Persian language as well as the indexed format to be useful for non-Persian researchers. The samples are generated from 3,539 multiple-choice questions with the least amount of annotator interventions in a way similar to the SciTail dataset. A designed multi-step process is adopted to ensure the quality of the dataset. The tagset consists of 3 common labels of contradiction, entailment and neutral. (Amirkhani, et, al. ,2020)

---

[4] https://www.nyu.edu/projects/bowman/multinli/
[5] https://github.com/dml-qom/FarsTail

## 3. The Proposed Systems

To the authors knowledge, the best results for contradiction detection systems are now obtained through machine learning and deep learning systems (such as Wang and Jiang, 2016), but in Persian and possibly in other low resource languages due to the lack of appropriate data with suitable volume for system training, these methods were not practically applicable till recent researches. Therefore, our first focus in this research is on introducing a practical approach based on general and specific rules in the Persian language. Secondly, for reaching better result for detection of some kinds of contradiction types that could not be achieved with rules, we created a medium size contradiction dataset through translation of existing English corpora (both manual and machine-translation methods). Then we train a Bert base system using this generated corpus.

Here we introduce our proposed system which consists of a rule based contradiction system and a Bert base deep learning one. Also we describe our baseline rule base system that is created for comparison to the main rule base system. This baseline system is based on a series of general features to identify semantic contradiction. The main rule base system, applies a data mining method (Frequent rule mining) on the development set, and automatically discovers the distinctive features of contradiction for the predefined contradiction categories. Details of these three systems are described in the following subsections.

### *3.1 The Basic Rule-based System*

In this section, we introduce our baseline rule-based system. So far, several rules and attributes such as the ones in (Marneffe et al., 2008) have been used to find contradiction. In our first proposed system, which we introduce as the baseline, we consider a wide range of syntactic and semantic features and use them to solve the problem of contradiction detection. In baseline, no new feature is introduced, we just integrated some feature values and used them as a comprehensive rule set. Before feature extraction, we apply a pre-processing procedure to the sentence pairs:

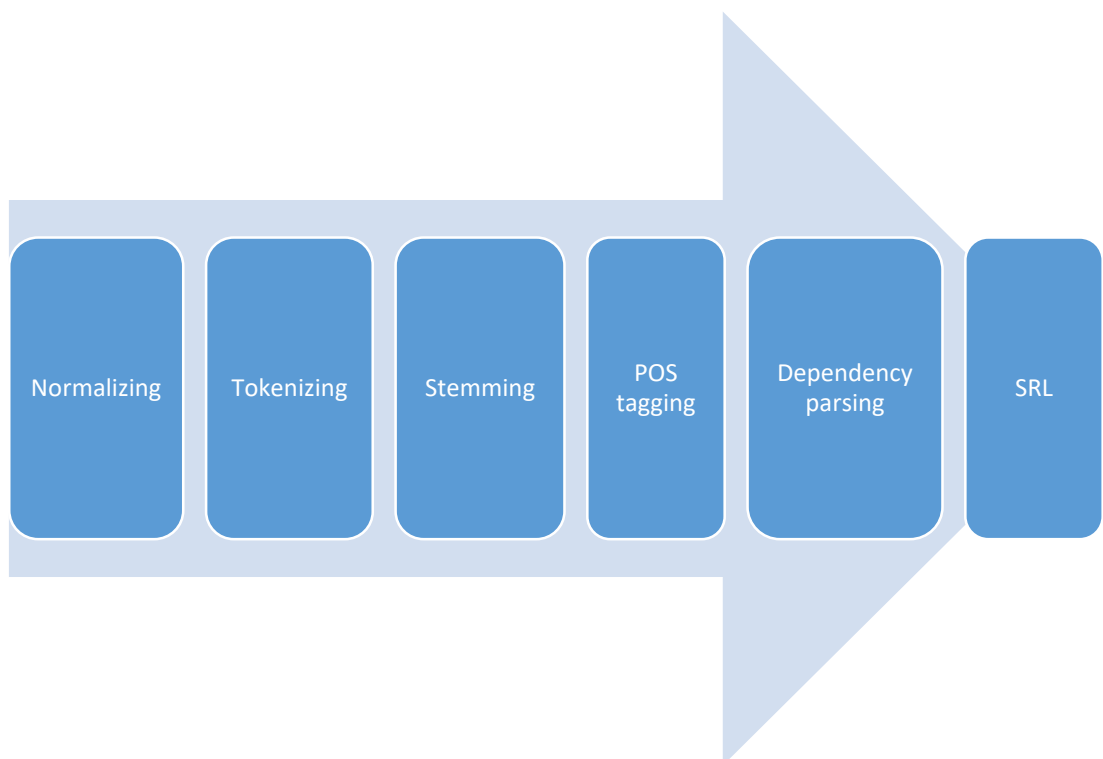

Fig. 1. Procedure of pre-processing

In the pre-processing procedure we used the following basic tools
- POS tagging models 100-tags that were trained over Peykareh[6] corpus with precision of 92%.
- We trained a CRF model on NER corpus of UT[7] for named entity recognition with accuracy of about 85% on 7 tags.
- We trained a dependency parsing model on Dadegan[8] corpus using Malt-parser with approximate precision of 85%.
- We used a SRL labeler tool, which was trained on Dadegan-SRL data with approximate precision of 75%.

The implemented features which are eventually used as rule set in baseline system are as follows:

---

[6] http://dbrg.ut.ac.ir/Bijankhan/
[7] http://www.parsigan.ir/datasources/NER/8
[8] http://www.peykaregan.ir/dataset/پیکره-وابستگی-نحوی-زبان-فارسی

1. Sentiment agreement: It has been proven by experience that many of the contradictory sentence pairs have different polarities, so the feature of the sentiment dis/agreement of two sentences is considered as one of the distinguishing features. For example, "john slept very well last night" and "john had a nightmare last night" have opposite sentiments. For sentiment analysis, a simple lexicon-based system based on sentistrenght[9] lexicons is implemented and used. The lexicon consists of almost 900 polar words. For calculating sentiment score, occurrence of polar words or their stems along with negation word list is considered.
2. Named entities comparison: It is observed that sometimes the existing named entities in contradictory sentences are different, in particular, the most famous ones are location, date or time. So this feature is also considered. E.g. "Mary went to Paris yesterday." And "Mary went to London yesterday". For this feature, after NE extraction, type and value of NEs are compared together for possible inconsistency.
3. Different size of the two sentences: Sometimes the length of the sentences of the text and the hypothesis is very different. This feature can sometimes be decisive for identifying contradiction.
4. Adjective similarity of two sentences: When the two sentences have the same syntactic structure, but they have different adjectives, such as the colors used in the sentence, it can be used to distinguish the contradiction. E.g. "the woman has a black shirt" and "the woman is wearing a blue shirt", For calculating this value, with observing ADJ part of speech tags in the both sentences, similarity or antonymy of two corresponding adjectives is investigated.
5. Verb similarity of two sentences :Verb is one of the most important elements of a sentence, so in many cases, the difference between verbs can well reflect the difference in the content of two contradictory sentences.
6. Negation :The negation in verbs with the same stem or negative adverbs can indicate the opposition in a pair of sentences. For example, "Ali went to school" and "Ali did not go to school."
7. Common words: The number of common words in two sentences is usually an important factor in determining the opposition or similarity of sentences. Especially when there is few other discriminative information.
8. Cosine similarity of two sentences: The use of the cosine similarity by removing the stop-words and normalization by the ratio of the length of two sentences can be an appropriate criterion for determining similarity or contradiction of a sentence pair. We use BOW vectors here. Certainly, the similarity value in entailments or similar sentences is more in comparison to contradictory pairs.
9. SRL argument similarity: In many contradictory cases e.g. structural contradictions, semantic arguments are used in different positions, so semantic tags can sometimes be a suitable attribute for identifying contradictions. E.g. "Water floats on oil" and "Oil is floating on water". In this system we investigate "TMP" and "LOC" labels.

---

[9] http://sentistrength.wlv.ac.uk/

10. Antonym: Obviously, the occurrence of conflicting words and phrases (antonyms) in similar positions in sentence pairs can indicate the contradiction in these sentences. Therefore, this feature is one of the important features used. E.g. "My clothes are still wet" and "My clothes are dry and warm".

In the proposed system, feature scores are normalized between 0 and 1 and then the weighted sum of these features are calculated for each sentence pair and converted to a contradiction score using some thresholds. These thresholds and weights are determined using a part of the dataset which is separated as development set with simulated annealing algorithm for optimization. Also we trained a machine learning system with different classifiers using this exact feature set for comparison. The details are stated in evaluation section.

### 3.2 Using Data Mining to Extract Rules for Main Rule-based System

As we stated in the introduction, the semantic contradiction includes different types, and if it is to identify contradictions with the rules, we should find and apply their own rules for each category. We introduce the second rule-based system in this section.

This system consists of two main parts. In the first section, using the associate rule mining method, the set of rules used to identify the contradictions, is automatically derived from considering the frequency of the occurrence of the rules in the development set. In the second part of procedure, these rules are selected and applied according to predefined categories. So we have different algorithms for different categories of contradiction sentences.

In the test phase, three modes may occur: First, ideally the type of input test data is specified, the second mode that we use a rule-based classifier to determine input category, and the third state that all the algorithms are applied to the input and the result is obtained by voting. Further details are given below.

A. For extracting the rules, the following steps are taken:
   1- Extraction of dependency relations, semantic role labels, sentiment analysis label and part of speech tags (POS) for both sentences
   2- Keeping dependency relations if:
      a. In which the common words of two sentences appear
      b. In which one side of an antonym relationship appears
      c. It has a certain tag like a "num".

   At this stage, these relationships need to be more general for frequent rule mining, so instead of putting the exact words inside the tuples, we put POS label of words. E.g. "amode(antonym-1, N-SING-COM)" or "num(NUM, N-SING-COM)" 3- Calculating the similarity of each two semantic arguments and put them in the form of (arg1, arg2, similarity interval). We use cosine similarity and defined similarity intervals by 0.3 steps. (0-0.0.3-0.6,0.6-1)
4- Determining sentiment analysis label for both sentences and put them in the form of (sentiment1, sentiment2). E.g. (Positive, Negative) or (Positive, Positive)
5- For each pair of verbs and quantifiers or negative conjunctions in the two sentences, check that verbs are positive or negative (based on POS tags) and make them in the form of tuples. For example, (Verb1-NEG, Verb2-POS) or (Verb1-NEG, Qunt1-NEG).
6- Now that we have gold label and some selected tuples for each sentence pair, Applying the associative rules mining using WEKA toolkit

The following flowchart indicate the procedure of frequent rules extraction.

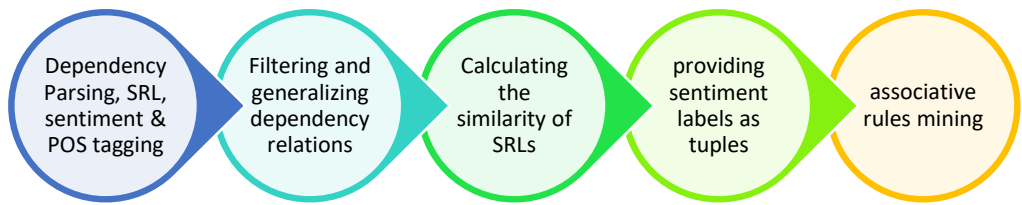

Fig. 2. Procedure of frequent rules extraction

Table 2. Sample of extracted rules

| Type | Rules | Example |
|---|---|---|
| Negation | 1. (Verb1-NEG, Verb2-POS) فعل جمله اول مثبت، فعل جمله دوم منفی (با بررسی هم ریشه بودن افعال)<br>2. (Verb1-NEG, Qunt1-NEG) (Verb1-NEG, Verb2-POS) فعل جمله اول منفی،سور در جمله اول منفی و فعل جمله دوم مثبت<br>3. (Verb1-NEG, Qunt1-NEG) (Verb1-NEG, Verb2-POS) (Verb1-NEG, Quant 2-POS) فعل جمله اول منفی،سور در جمله اول منفی و فعل و سور جمله دوم مثبت<br>4. (Verb1-NEG, ADV1-NEG) (Verb1-NEG, Verb2-POS) فعل منفی و قید منفی در جمله اول و فعل مثبت در جمله دوم | ۱. علی به مدرسه رفت. ( Ali went to school.)#علی به مدرسه نرفت. ( Ali did not go to school)<br>۲. هیچ کس به مدرسه نرفت. ( no one went to school) # علی به مدرسه رفت. ( Ali went to school)<br>۳. هیچ کس به مهمانی نیامد . ( no one came to the party) # همه به مهمانی آمدند. ( everybody came to the party)<br>۴. هرگز به تهران برنمی گردم. ( I never come back to Tehran) # به تهران برگشتم. ( I came back to Tehran) |
| Numeric | 1. Num(NUM1, N-SING-COM2) Num(NUM2, N-SING-COM2) [and basically comparison the args] وجود عدد و اسم در هر دو جمله و مقایسه آرگومانها<br>2. Num(NUM1, antonym1) Num(NUM2,antonym2) وجود عدد در رابطه با دو کلمه متضاد در دو جمله | ۱. سه دختر در خیابان نشسته بودند. ( 3 girls were sitting in the street) # ۵ دختر در خیابان بودند. ( 5 girls were in the street)<br>۲. سه دختر در مدرسه نشسته بودند. ( 3 girls were sitting at school) # سه پسر در کلاس درس می خواندند. ( 3 boys were studying in the classroom) |
| Structural | (A0,A1,E2) | آب روی روغن شناور می ماند . ( Water floats on oil) # روغن روی آب شناور است. ( Oil is floating on water) |

B. In the next step, the extracted frequent rules are assigned to different predefined contradiction categories manually.  As discussed in the introduction section, seven classes are considered for contradiction types including Antonym, Negation, Numerical, Structural, Factive, Lexical and World Knowledge. Due to the complexity of the last three ones we merged these three categories under "others" category. So we have five predefined categories of numeric, negation, structural, anonym and others. In rule base system we have implemented extracted rules for the 4 first categories and use our baseline system for "others" category. Table 2 shows some of the rules extracted for the predefined categories.

To distinguish this system from baseline rule base system, we call it DM-rule base system.

### *3.3 Bert base deep learning system using generated data*

As we stated at the beginning of section 3, due to the emergence of Bert[1] models (Delvin, et al., 2018) and their positive performance in similar tasks, we decide to fine tune the Persian pre-trained language model -ParsBert- (Farahani, et al., 2020) on our translated data as well. As training data, we machine translated part of SNLI and MultiNLI datasets in addition to nearly small part of manual translated data and Farstail. Our results and implementation details on this deep system are presented in section 4.3.

For fine-tuning we used (Gao et al., 2021) implementation[1] which is a PyTorch solution of natural language inference (NLI) model based on Transformers.

## 4. Evaluation

### *4.1 Datasets*

In this study, we prepared three datasets. (1) 1000 sentence pairs manual translated from SNLI corpus of Stanford University, of which 324 are contradictory and 676 are neutral and entailment. (2) Set of 250 sentence pairs (130 contradictory and 120 neutral sentence pairs) containing different categories of contradictions that were manually generated for testing the system. Also (3) we machine translated part of SNLI and MultiNLI dataset using Google Translate for training our deep learning system.  Our gold datasets statistics and sample distribution is presented in Table 3 (Dataset1 and Dataset2) and details of our automatic generated dataset is indicated in table 4.

---

[1] BERT: Pre-training of Deep Bidirectional Transformers for Language
[1] https://github.com/yg211/bert_nli

Contradiction detection in Persian text    17Table 3. Gold datasets statistics and sample distribution

| Dataset | type | Sentence pairs | Contradictory pairs |
|---|---|---|---|
| Dataset1 | Manual Translated SNLI | 1000 | 325 |
| Dataset2 | Negation | 42 | 26 |
| | Numeric | 47 | 28 |
| | Antonym | 52 | 26 |
| | Structural | 31 | 15 |
| | WK+Factive+Lexical | 78 | 35 |
| | SUM | 250 | 130 |

Table 4. Description of our machine translated data (Dataset3)

| Data source | #of sentence pairs | Contradictory pairs |
|---|---|---|
| SNLI-test | 10000 | 3333 |
| SNLI-dev | 10000 | 3333 |
| MNLI-dev | 10000 | 3333 |
| All automatically translated data | 30000 | 10000 |

### *4.2 Metrics*

Evaluation measurements here are similar to most of NLP tasks and consist of precision, recall, and F-measure, which are defined as follows:

$$\text{Precision}(P) = \frac{\text{correct system decisions}}{\text{all system decisions}} \quad (1)$$

$$\text{Recall}(R) = \frac{\text{correct system decisions}}{\text{what system should have decided}} \quad (2)$$

$$F - measure = \frac{2 \cdot P \cdot R}{(P + R)} \quad (3)$$

### *4.3 Results and Discussion*

The results of presented systems evaluation are as follows. First of all, evaluation of the baseline system is reported. Also as we mentioned, we trained a machine learning system using extracted features in section 3.1 to compare with our baseline. In machine learning system, dataset1 is used for training (668 samples out of 1000 to balance the classes) the system and different classifiers is tested in Weka[1] toolkit. Among tested classifiers the best result obtained through naïve Bayes, RBF network and Meta.Multi.class.Classifiers.

Table 5. baseline evaluation on the two gold datasets

| System name | Train/Development set | Test set | Precision | Recall | F-measure |
|---|---|---|---|---|---|
| Baseline rule-based | Dataset1 | Dataset2 | 62 | 68.2 | 65.1 |
| Baseline rule-based | Dataset2 | Dataset1 | 53.7 | 86.4 | 66.2 |
| ML-naïve Bayes | Dataset1 | Dataset2 | 55.5 | 72 | 62.9 |
| ML-RBF networks | Dataset1 | Dataset2 | 56 | 65.5 | 60.4 |
| ML-Meta.Multi.class.Classifier | Dataset1 | Dataset2 | 60.3 | 64 | 62.1 |

Secondly, evaluation of proposed rule base system (DM-rule base) is presented. As we mentioned, we have provided test samples for each contradiction category. In this section, we apply each algorithm (rule matching) to its associated samples. The results obtained are as follows in table 6.

Also we tested the system for dataset without considering indicated labels (such as negation, numeric, …) and applied all algorithms for each sentence pairs. The results are indicated in the last row of table 6. The results show that our rules are sufficiently discriminative for unlabelled data (having just contradiction or not contradiction tags).

Table 6. Evaluation of separate algorithms for each category (DM-rule base)

| category | Precision | Recall | F-measure |
|---|---|---|---|
| Negation | 92.8 | 85.7 | 89.1 |
| Numeric | 89.3 | 82.14 | 85.6 |
| Antonym | 86.5 | 84 | 85.2 |
| Structural | 67 | 46 | 54.5 |
| WK+Factive+Lexical | 62 | 68.2 | 65.1 |
| Average | 79.52 | 73.21 | 75.9 |
| Algorithm applied on dataset without considering indicated labels | 73.2 | 72.5 | 72.9 |

---

[1] https://www.cs.waikato.ac.nz/ml/weka/s

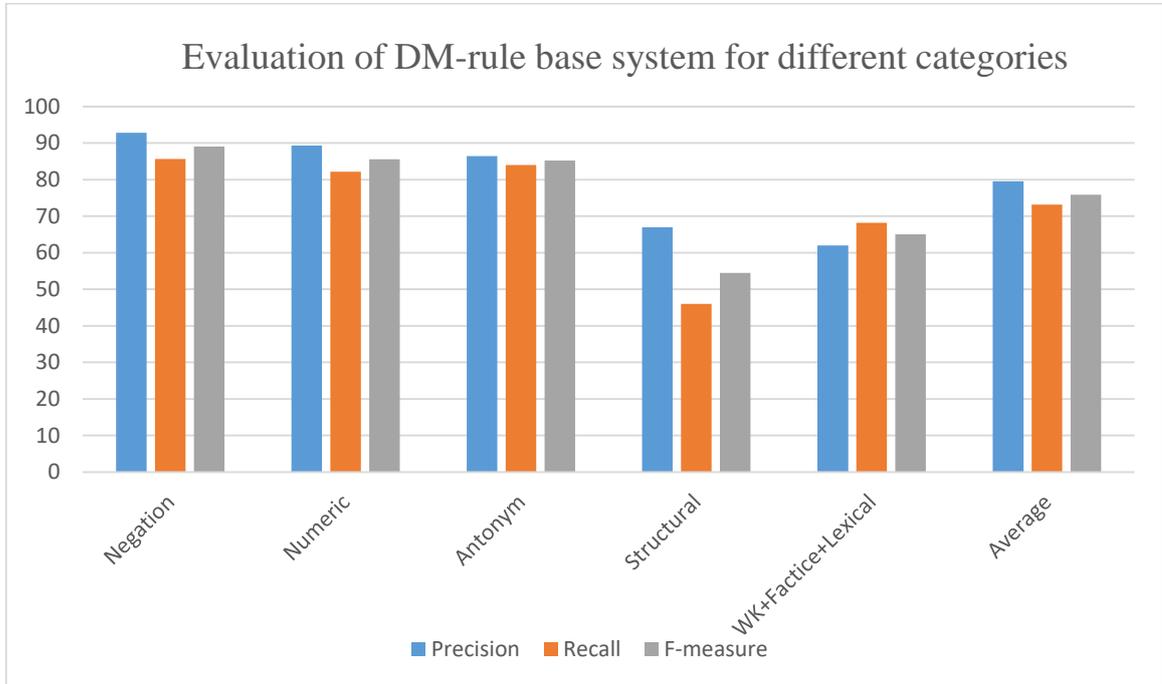

Fig. 3. evaluation of separate algorithms (rule matching) for each category

As it can be seen in table 6, DM-rule base system for the first categories (Negation, Numeric and Antonym) has very good performance. This is because of the fact that in development dataset there are proper examples of these types, so their rules are well extracted, on the other hand sentence pairs belonged to these categories are not too complicated mostly.

For structural category, the performance is not promising due to the lack of a semantic role labelling tool with the proper accuracy in the Persian language. For "others" category, we couldn't extract proper specific rules and just tested the samples using our baseline system. The results are not very good due to both sentences complexity and lack of using knowledge resources. Although overall performance of the proposed system is promising.

As the results of DM-rule base system shows, System performance is not good for some categories. (E.g. WK or factive), so as in section 3 stated, a Bert-based deep learning module was introduced to improve overall system performance. For training and evaluation this module, we use variation of datasets. We trained this module on our translated dataset (Dataset3) and Farstail, and evaluate them in two ways: once each one on its own test datasets and once on Dataset1(for fair comparison to rule base systems). Description of our machine translated data is shown in table 4. Also as indicated in related datasets in section 2, Farstail is consists of almost 10000 sentence pairs (7000 for training set and 1500 for each of test and development set). The evaluation results of Bert base deep learning system for "contradiction" class are indicated in Table 7 and 8.

Table 7. Evaluation of Bert base deep system on Dataset3 and Farstail with their own test dataset for "contradiction" class

| Train dataset | Test dataset | P | R | F |
| --- | --- | --- | --- | --- |
| Dataset3 | 3k separated from dataset3 | 71 | 75 | 73 |
| Farstail train | Farstail test | 73 | 67 | 70 |
| Dataset3+ Farstail | 3k separated +Farstail test | 79 | 72 | 75 |

Table 8. Evaluation of Bert base deep system on Dataset1 for "contradiction" class

| Train dataset | Test dataset | P | R | F |
| --- | --- | --- | --- | --- |
| Dataset3 | Dataset1 | 66 | 75 | 70.2 |
| FarsTail train | Dataset1 | 51 | 62 | 56 |
| Dataset3 + FarsTail | Dataset1 | 69 | 71.2 | 69.8 |

Among Persian similar researches, (Khodadadi et, al.,2015) detects contradiction relation in one sentence. But we tested deep learning algorithm of (Wang and Jiang, 2016)[1] which reported the highest results on SNLI dataset to the best of our knowledge, with our dataset too. In table 6 performance of 4 algorithms of 1) baseline 2) best machine learning system, 3) Bert base deep learning system and 4) DM-rule-based system are compared.

Table 9. Performance evaluation of all implemented systems

| System name | Precision | Recall | F-measure |
| --- | --- | --- | --- |
| Baseline rule-based | 53.7 | 86.4 | 66.2 |
| ML-naïve Bayes | 55.5 | 72 | 62.9 |
| DM-rule-based | 79.52 | 73.21 | 75.9 |
| Deep learning (Wang and Jiang, 2016) | 39.4 | 41 | 40.18 |
| ParsBert Deep learning | 69 | 71.2 | 69.8 |

As the result indicates, the best total performance is obtained through Bert base deep learning system. But it still is lower than DM-rule base performance for 3 categories of Negation, Numerical and Antonymy (F-measure=87). So we combine our two system of DM-rule base (to use for three mentioned categories) and Bert base deep system for others. Our final total performance is about 80 for all categories.

---

[1]   https://github.com/codedecode/Recognizing-Textual-Entailment

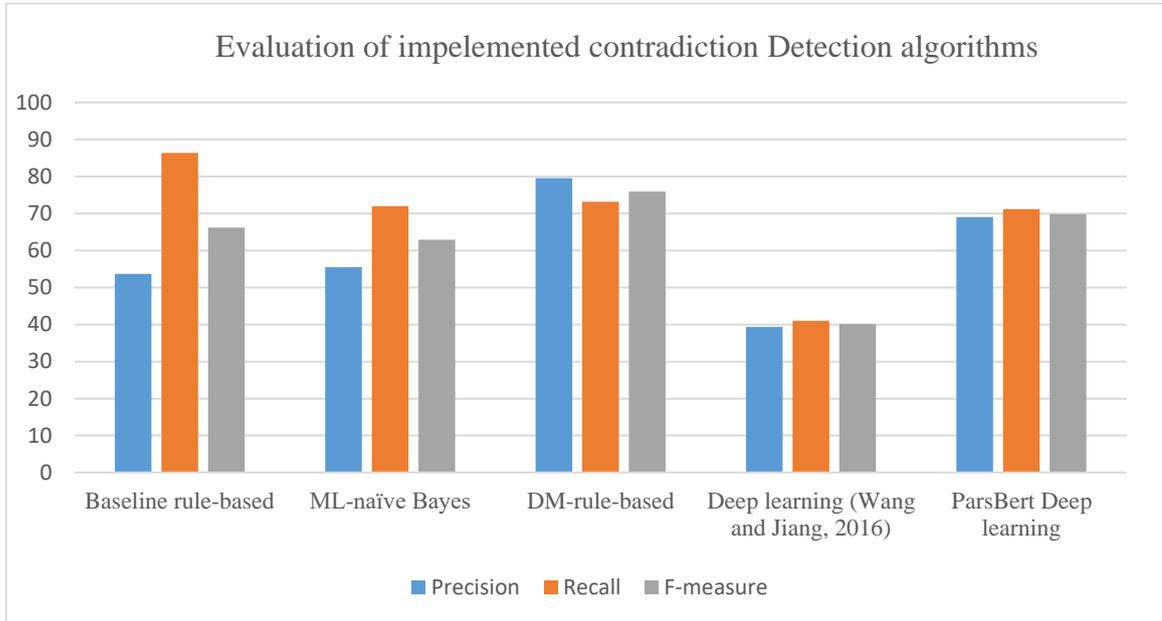

Fig. 4. Evaluation of implemented algorithms

## 5. Conclusion

In this research, two rule-based systems and a Bert base deep learning one are introduced to explore the semantic contradiction in Persian sentences. These systems are compared to various machine learning and deep learning methods on Persian text and performed better. The baseline system is based on a series of general features to identify semantic contradiction, but the DM-rule base system, using a development set, automatically discovers the distinctive features of contradiction (especially for Persian texts) for the seven categories. In this regard, the frequent rule mining method is used and the implementation of frequent rules assigned to each category leaded to promising results for some categories. Considering the fact that for low resource languages such as Persian language, due to lack of appropriate datasets, machine learning and deep learning systems may not be effective, the development of rule-based methods with proper functioning can help to identify semantic contradictions. As the proposed system has a good performance comparable to best systems in the world. In the case of combinational category, due to the complexity of sentences and lack of utilization of appropriate knowledge resources, good performance was not achieved. So a Bert base deep system is presented to cover the weakness for other categories. Our hybrid system has performance of 73 (F-measure) for contradiction detection in Persian text.

As a future work, we plan to create and use a contradiction embedding model for possible better performance.

22  Z. Rahimi and M. ShamsFard

.